\documentclass[11pt]{article}

\usepackage[letterpaper,margin=1in]{geometry}
\usepackage{amsmath}
\usepackage{booktabs}
\usepackage{caption}
\usepackage{enumitem}
\usepackage{graphicx}
\usepackage{hyperref}
\usepackage{natbib}
\usepackage{xcolor}
\usepackage{xurl}
\usepackage{array}
\usepackage{longtable}
\usepackage{pdflscape}
\usepackage{placeins}
\usepackage{svg}
\usepackage{float}
\usepackage{placeins}
\usepackage{flafter}
\usepackage{fvextra}
\usepackage{subcaption}
\usepackage{graphicx}
\usepackage{wrapfig}
\usepackage{listings}

\hypersetup{
  colorlinks=true,
  linkcolor=blue,
  citecolor=blue,
  urlcolor=blue
}

\setlist[itemize]{leftmargin=1.5em}
\setlist[enumerate]{leftmargin=1.5em}
\title{LifeVal: Evaluating AI Model Performance on Personalized Agentic Consumer Tasks}
\author{%
  \parbox{0.9\textwidth}{%
    \centering
    Jenny Yang, Elizabeth Proehl, Anjan Bharadwaj, Emily Fan,
    Charlotte Cole, James Kim, Noam Brown, Phoebe Thacker,
    Amelia Glaese, Tejal Patwardhan\\[0.5em]
    OpenAI
  }%
}
\begin{document}

\maketitle

\begin{abstract}
We introduce LifeVal, a benchmark evaluating AI model performance on personalized agentic consumer tasks. LifeVal covers the most valuable life tasks that a personalized digital superassistant could automate, representing over \$1.7 trillion USD in annual willingness-to-pay across 148 consumer task categories (e.g., managing finances, moving apartments, searching for a job, finding childcare). To construct our activity distribution, we conducted a new survey of willingness-to-pay for a digital superassistant across a demographically representative sample of over 3,800 consumers, cross-referencing our results against the American Time Use Survey and the Multinational Time Use Study \citep{bls2024atus, mtus2020}. For the most valuable consumer activities, we recruited task creators to create AI evaluation tasks personalized using real ChatGPT conversations and memories, reference files, and personal context. Task creators had to pass our expertise threshold (including personally completing the task multiple times and seeing the results through in their real lives for over a year) and also write a rubric, ensuring that our evaluation measured the best personalized completion for that task. All tasks had to pass at least three rounds of human quality-control before acceptance into our final evaluation sets. 
\end{abstract}
\begin{figure}[H]
    \centering
    \includegraphics[width=1\linewidth]{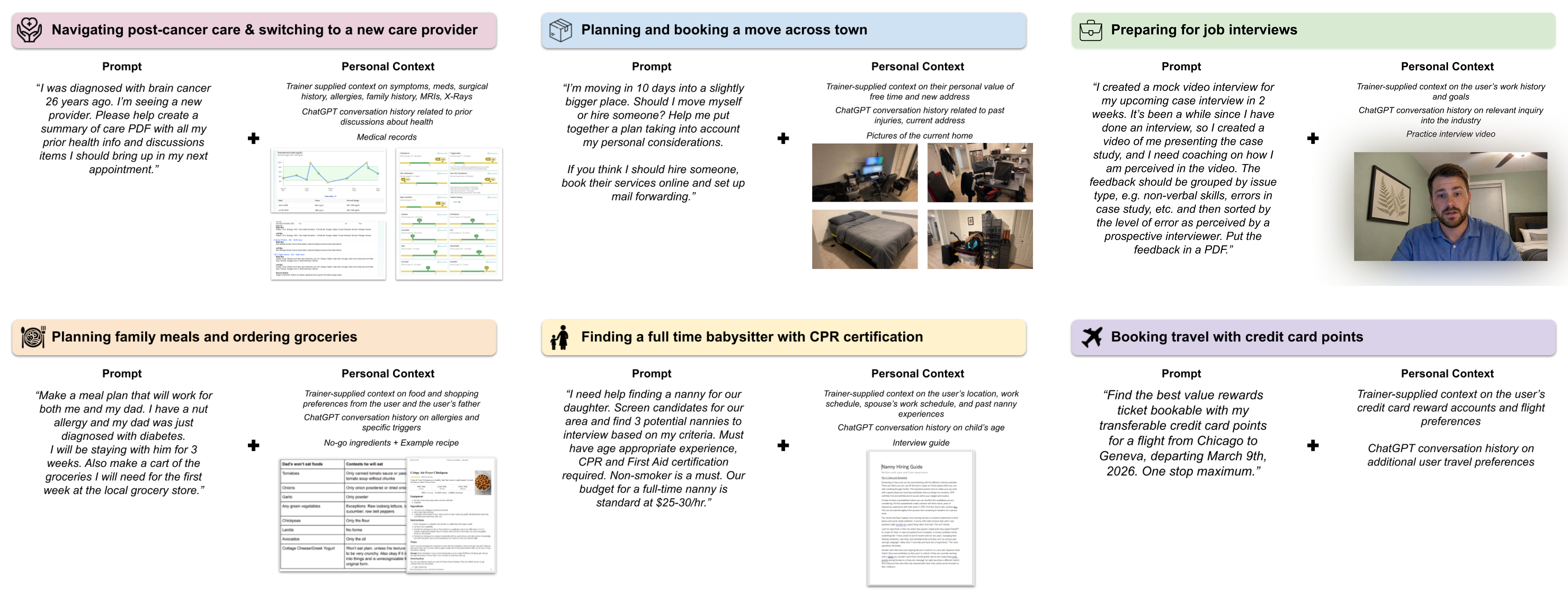}
    \caption{Example LifeVal Tasks. Any personal information in the paper, including any images with personal information, has been deidentified or synthetically generated to protect our task creators' privacy.}
    \label{fig:task_examples}
\end{figure}

\section{Introduction}
AI models are increasingly acting as digital superassistants: highly capable, personalized agents with deep integration across users’ digital lives. To understand and forecast the impacts of such models, we need to measure their capabilities. One way to measure the consumer capabilities of such models is through usage metrics. However, while usage data tells us about tasks already requested by users, adoption often lags capabilities, and usage metrics don't capture newer capabilities that take time to adopt, or tasks that users don't think to ask \citep{patwardhan2025gdpval}. To better measure the capabilities of AI models in consumer settings, we introduce LifeVal, a benchmark explicitly designed to measure AI capabilities on real-world tasks that consumers deeply value.

LifeVal tasks capture valuable, long-horizon consumer needs:

\begin{itemize}
    \item \textbf{High Value:} LifeVal tasks are pulled from the highest-value tasks from a demographically-representative survey of willingness to pay for AI superassistance, triangulated against the American Time Use Survey and Multinational Time Use Study, which compiles comparable time-use surveys from 25 countries \citep{bls2024atus, mtus2020} \footnote{Australia, Austria, Belgium, Bulgaria, Canada, the Czech Republic, Denmark, Finland, France, Germany, Hungary, Israel, Italy, the Netherlands, Norway, Peru, Poland, Serbia, Slovenia, South Africa, South Korea, Spain, Sweden, the United Kingdom, and the United States.}. 
    \item \textbf{High Context:} LifeVal tasks require the model to reference detailed personal data, lengthy ChatGPT conversation histories, and context across different applications. 
    \item \textbf{Personalized:} LifeVal tasks are highly personalized and subjective, requiring models to adapt to individual preferences. 
    \item \textbf{Agentic:} LifeVal tasks require taking complex actions on a computer, including interacting with websites, accessing account states, filling forms, producing artifacts, and managing digital transactions. 
    \item \textbf{Objectively gradable:} Despite their subjective nature, LifeVal tasks are possible to objectively grade when conditioned on the provided context.
\end{itemize}

\section{Related Work}
LifeVal sits at the intersection of three literatures that have mostly remained separate: agentic benchmarks, personalization and memory benchmarks, and economic literature on task augmentation and automation. 

AI evaluations are shifting from single-turn academic benchmarks toward interactive, agentic tasks that test whether models can reliably complete real workflows \citep{osworld2024, webarena2024}. Performance has risen rapidly since early baselines, including WebArena’s original 14\% vs.\ 78\% human gap, OSWorld’s original 12.24\% agent performance vs. 72.36\% human  \citep{webarena2024, osworld2024}. More recent verified variants report much stronger results, including 67.3\% on WebArena-Verified and 75.0\% on OSWorld-Verified for GPT-5.4 \citep{openai_gpt54_2026}. This literature suggests several design lessons for LifeVal: (i) partially subjective outcomes are often evaluated with rubrics \citep{webvoyager2024}; (ii) reinforcement learning environments improve reproducibility relative to the live web \citep{webarena2024, osworldverified2025}; (iii) evaluation harnesses must be hardened against reward hacking \citep{rdi_benchmark_hacking_2026}.

Related work examines how AI systems infer, retain, and apply user preferences \citep{surveypersonalization2025}. PrefEval \citep{zhao2025prefeval} evaluates whether models can remember preferences over long synthetic conversations, while other benchmarks study personalized assistance in conversational and research settings \citep{permabenchmarking2026, personalizeddeepresearch2026}. More closely related to LifeVal, PersonalWAB \citep{cai2025personalwab} combines real shopping histories with generated user profiles to evaluate personalized search and recommendation performance, and Persona2Web \citep{kim2026persona2web} evaluates whether agents can infer preferences from synthetic user profiles and browsing histories to tasks on the open web. Our work extends personalized agent evaluation by combining real ChatGPT conversation histories and memories with tasks requiring browser and computer use across everyday domains. The same users contribute task-specific rubrics, providing ground truth about whether the agent successfully used their personal context. 

This paper is also closely related to a large literature in economics on task-automating or augmenting technologies \citep{autor2003skill, acemoglu2011skills, acemoglu2019automation}. When technology sharply reduces the cost of a task, households gain purchasing power, time, or both, and can reallocate those resources toward other goods and services, especially complements to the now-cheaper task \citep{greenwood2005engines}.

More broadly, automation can create new consumer demand rather than merely substitute for labor. Agricultural mechanization and productivity growth reduced food prices, allowing households to devote less spending to food and more to manufactured goods and services \citep{gollin2002role, herrendorf2013two}. Similarly, electrification and labor-saving appliances reduced the time required for cooking, cleaning, and laundry, shifting time out of unpaid home production and into market work and other activities \citep{greenwood2005engines, ramey2009time}.

When the marginal cost of completing a task falls far enough, technology can also expand access. In the U.S., digital retail increased consumer surplus by adding variety and convenience, especially for households outside dense urban markets \citep{brynjolfsson2003consumer, sinai2004geography, dolfen2023assessing}. Such effects may be larger when a technology lowers the cost of many consumer tasks at once.

Many LifeVal tasks — coordinating home services, managing finances, communicating with providers, and navigating taxes — are inputs into household production rather than final consumption. These activities are quantitatively important, as households devote substantial time to home production and to purchasing and coordinating goods and services \citep{aguiar2007measuring, ramey2009time}. Tax filing is one clear example, where many low-income households pay for assistance to access tax-administered benefits, and filing frictions affect claiming behavior \citep{bhargava2015psychological}. In most other domains, however, the relevant margin is the cost of administrative burden. Incomplete take-up of social services reflects learning costs, paperwork requirements, and other barriers that deter eligible households \citep{currie2004take, finkelstein2019take, herd2019administrative}. Reducing the cost of these tasks may therefore generate welfare gains through time savings, lower expenditures, and expanded access for households that would otherwise be unable to participate.

\section{Task and Benchmark Construction}
\label{sec:construction}

\subsection{Prioritizing consumer life tasks}

To construct the initial set of tasks for LifeVal, we used data from the American Time Use Survey (ATUS, 2024; n=7,700) and the Multinational Time Use Study (MTUS, 2020; n=8,782). ATUS is a detailed time-use taxonomy created by the U.S. Bureau of Labor Statistics (BLS) that documents how people allocate their time across daily activities. MTUS is a database of international time-use diary surveys, designed to make activity categories comparable across countries. Together, these datasets provide a detailed view of how people spend their time across a full 24-hour period, including differences between weekdays and weekends and across different times of year. We cleaned the data based on the procedure outlined in \ref{app:datacleaning}.

We then estimated the potential value of an AI digital superassistant for each task. We considered whether task categories could simply be weighted by the amount of time people spend on them. But for consumer life tasks, time spent is often a poor proxy for value: some tasks are relatively infrequent or short in duration, yet highly consequential when they arise, such as tax preparation, choosing health insurance, or obtaining preventive health care. Moreover, sometimes life tasks that take a lot of time (e.g., sleeping, eating) can only be moderately impacted by a personal digital superassistant; an AI model can give advice about these activities, but it cannot sleep for you or eat for you. This makes measuring AI impacts on life tasks different from work-related tasks. For work tasks, the value from an AI superassistant can often be approximated more directly using average wages multiplied by hours worked for the relevant occupation. 

Because no comparable market wage exists for many personal life tasks, we supplemented the ATUS/MTUS task data with a willingness-to-pay survey (n=3,837), with recruitment targeting US Census distributions for age, gender, and ethnicity from Census Bureau population estimates (2021), and income from the Current Population Survey Annual Social and Economic Supplement (CPS-ASEC, 2025) \citep{census2021population, cpsasec2025}. We specified a hypothetical perfect digital superassistant that could provide advice and perform digital tasks, but could not perform physical work. Respondents estimated the maximum amount they would have paid over the course of 2025 for digital assistance with each activity (respondents remained responsible for the cost of any goods or services arranged). Each respondent evaluated a randomized subset of 30 activity categories.

The resulting data informed a prioritized distribution of 148 activity categories derived from the most detailed level of the ATUS classification, used to guide further LifeVal task collection. Summing the estimated U.S. annual willingness to pay across these categories yields approximately \$1.7 trillion (Table~\ref{tab:atus-categories}). We compared these valuations with external proxies for activity importance, including BLS consumer expenditure, occupational wages, and ATUS wellbeing measures. The valuation question is provided in \ref{app:survey}, aggregate willingness-to-pay estimates by broad activity category in \ref{app:task_categories}, and a discussion of the external comparisons in \ref{app:surveyvalidation}. The resulting tasks span the most important areas that people want help with, including caregiving, health, education, household administration, personal finances, shopping, and accessing government services.

\subsection{Task writer recruitment, consent, and personalization data}

\begin{figure}[H]
    \centering
    \includegraphics[width=0.75\linewidth]{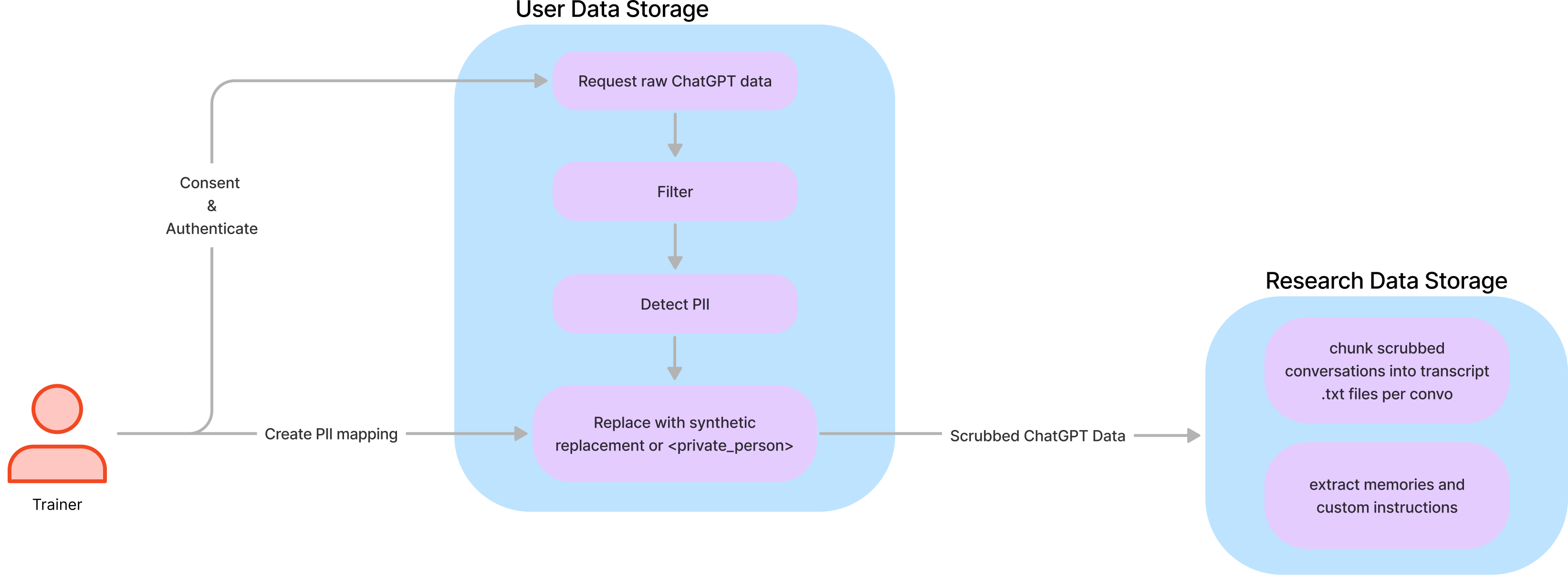}
    \caption{Task writer personalization data processing}
    \label{fig:personalization}
\end{figure}

Our belief is that only an ``expert'' in the task knows what the truly correct outcome is in the long run \textit{for them}. In LifeVal, we did not want generic tasks; each task instead represents what is useful and correct for the writer. We personally reviewed every task proposed by expert task writers to confirm their experience met the expertise thresholds for the project, including back-and-forths with additional information requests, and conducted live virtual trainings and office hours. When writing tasks, expert task writers replaced personally identifiable information (PII) with synthetic data and consented to share their ChatGPT data which itself was de-identified before research use, as shown in figure \ref{fig:personalization}.

\subsection{Task Creation}

Each task consists of the following components:

\begin{itemize}
    \item \textbf{Request (Prompt):} Clearly specifies what the task writer wants the model to accomplish. 
    \item \textbf{Personalized Context:} Provided to enable accurate personalized task completion. While many personal queries seem highly subjective, providing detailed personal context significantly reduces ambiguity. For example, the general question ``What should I have for dinner?'' becomes objectively answerable if the model has access to the user's preferences and recent dining history. Personalized context includes: 
    \begin{itemize}
        \item \textbf{ChatGPT Conversations and Memories:} Relevant excerpts from prior conversations and memories. At evaluation time, the model sees their entire conversation dump (minus any conversations they archived or deleted). 
        \item \textbf{Reference Files:} Any additional context provided to complete the task in file form, e.g. medical test results, last semester's course schedule. 
        \item \textbf{Additional Task Context:} Any other essential details or resources that the model would need to successfully perform the task that are not available in ChatGPT data, reference files, or web browsing. 
    \end{itemize}
    \item \textbf{Rubric:} To ensure objective grading despite the inherent subjectivity of personalized tasks, writers also create rubrics for each task.
    \item \textbf{Reinforcement Learning Environment (RLE):} Each task has corresponding environment(s) where the model can complete the agentic actions required by the task. We also give the model a browser tool to search an offline version of the web. 
    \item \textbf{Starting State:} Initial data or information required within the environment to complete tasks (e.g., an email in the inbox containing medical results).
\end{itemize}

\begin{figure}[H]
    \centering
    \begin{minipage}{0.48\textwidth}
        \centering
        \includegraphics[width=\linewidth]{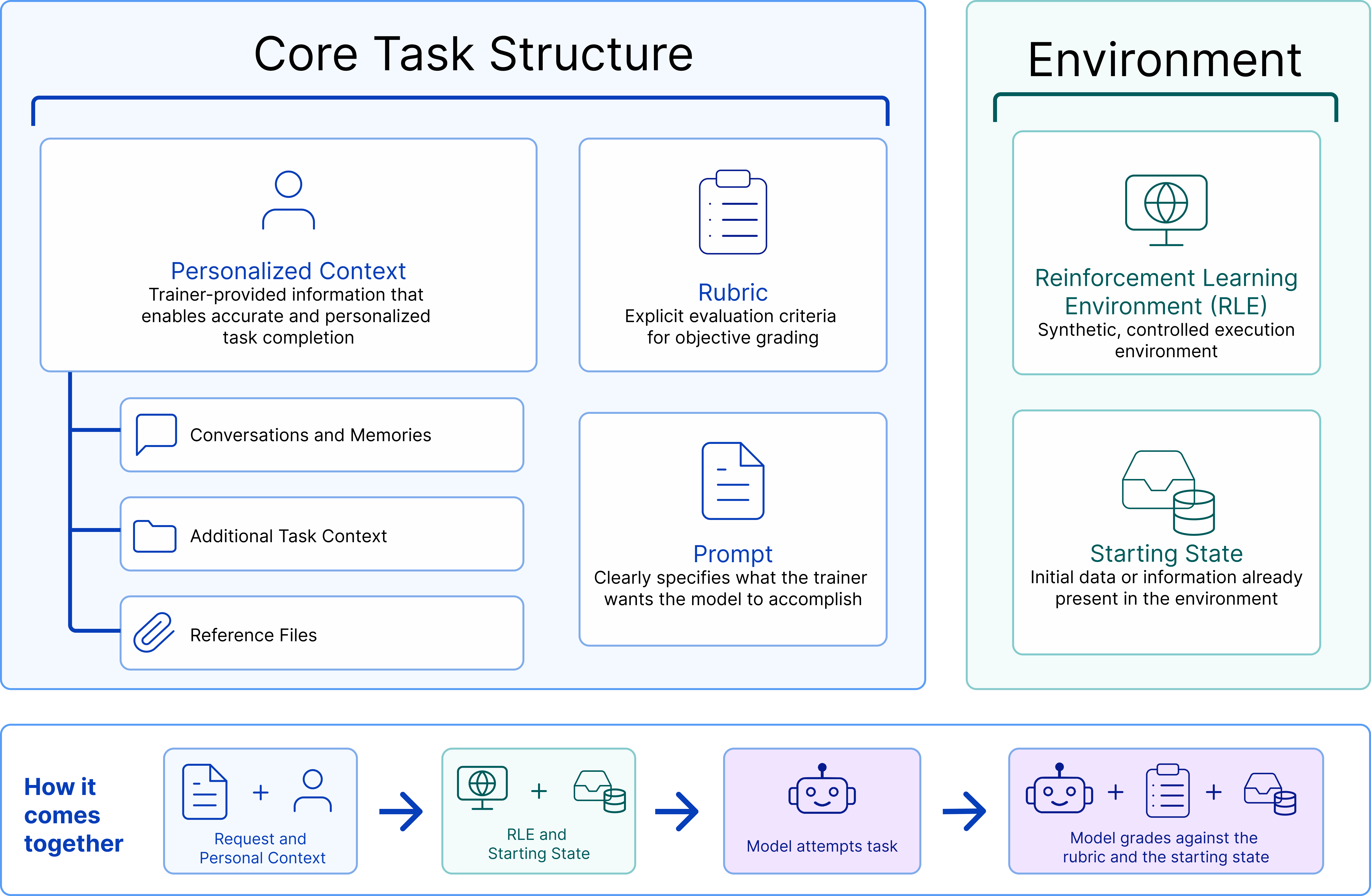}
        \caption{Task structure}
        \label{fig:task_structure}
    \end{minipage}
    \begin{minipage}{0.48\textwidth}
        \centering
        \includegraphics[width=\linewidth]{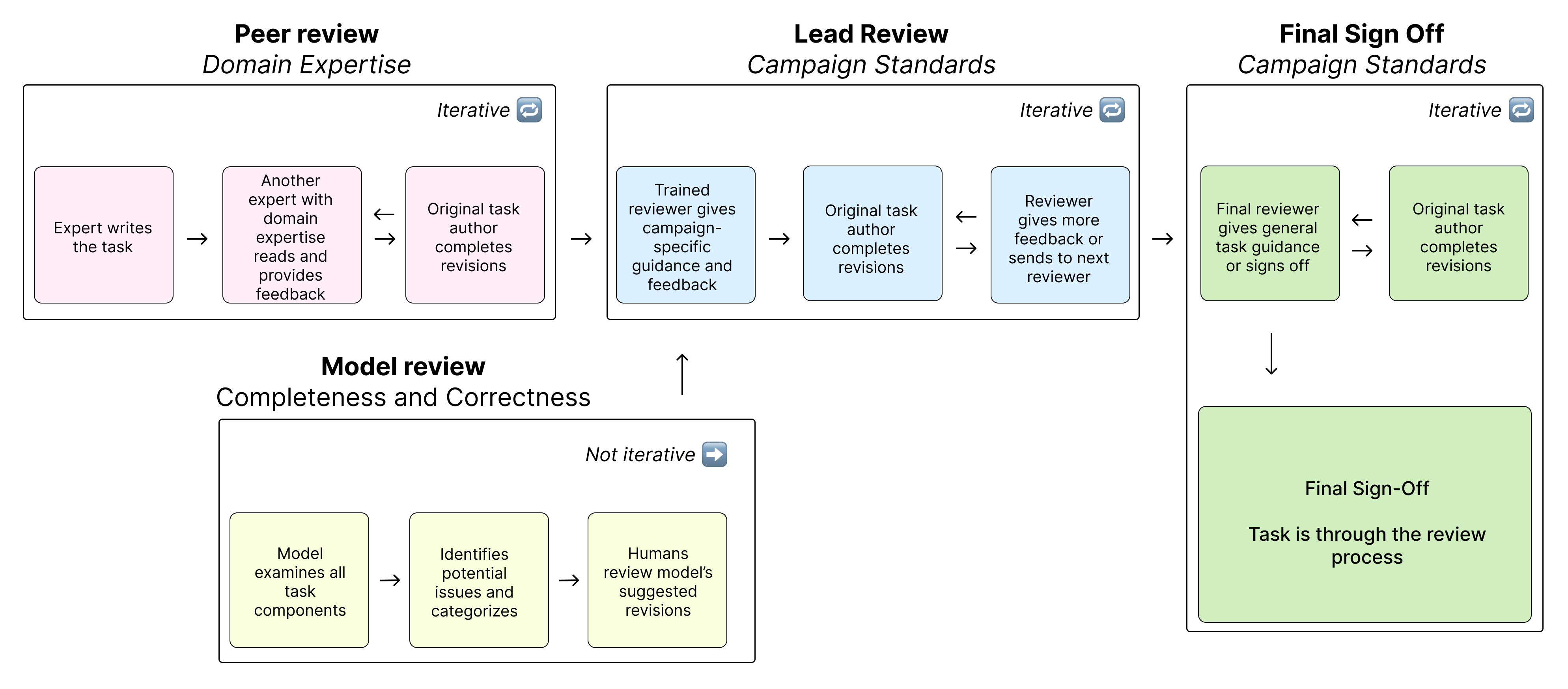}
        \caption{Task quality control loop}
        \label{fig:quality_control}
    \end{minipage}
    \hfill
\end{figure}

\subsection{Quality control for tasks, environments, and rubrics}
We employ rigorous quality control (QC) procedures, involving both human and automated model review to ensure task accuracy, realism, and fairness.

\subsubsection{Human QC}

LifeVal tasks underwent at least three rounds of human review:

\begin{enumerate}
    \item \textbf{Peer Review:} Conducted by another task writer with domain expertise relevant to the task if needed. The peer reviewer assessed task accuracy, realism, and completeness from the perspective of practical domain knowledge. 
    \item \textbf{Lead Review:} A trained reviewer evaluated the task against campaign standards, including whether the task demonstrated sufficient difficulty for the campaign and was fair i.e., possible to complete with the provided information. 
    \item \textbf{Final Sign Off:} A third expert reviewer reviewed the completed task and signed off when it was ready for inclusion.
\end{enumerate}

\subsubsection{Model QC}

In parallel with human review, model-driven QC helped ensure that the rubric criteria were fair and solvable given the provided inputs:

\begin{itemize}
    \item A model is given all task components, including the prompt, personal context, and environment, and attempts to complete the task. 
    \item Another model reviews the completion of the task and categorizes failures into rubric, prompt/context, environment, or model-related problems. 
    \item Based on the identified issues, the model proposes targeted corrections.
    \item   Human reviewers assess the model's proposed fixes and decide whether or not to implement them. 
\end{itemize}

\subsection{Grading}

We use two automated graders, both of which use GPT-5.6. 
\begin{itemize}
    \item \textbf{Primary Grader:} The purpose of the primary grader is to assess whether the task was completed correctly. The grader receives the original task prompt, relevant conversation history and memories, task rubric items and their available points, the model’s final response, any files or artifacts the model created, and a backend state diff showing the change in the task RLE from start to finish. The grader is instructed to score each rubric item independently using only the provided evidence. It must verify that any artifacts the model claims to have created exist, consider backend state changes for tasks involving real side effects, and only give credit for actions that were actually completed. 
    \item \textbf{Hack Grader:} The goal of the hack grader is to detect successful environment hacking. We consider any instance when the model accesses internal RLE APIs to be hacking. This is discouraged because real sites are unlikely to have these types of open endpoints for a model to use. If hacking is detected, the sample receives a score of 0. For additional information about the grading setup, see \ref{app:grading}.
\end{itemize}

\section{Results}
\label{sec:results}
We created a set of 169 tasks and evaluate GPT-5.2, 5.4, 5.5, 5.6, and GPT-6 Astra. 

\begin{figure}[ht]
    \centering

    \begin{subfigure}{0.48\textwidth}
        \centering
        \includegraphics[width=\linewidth]{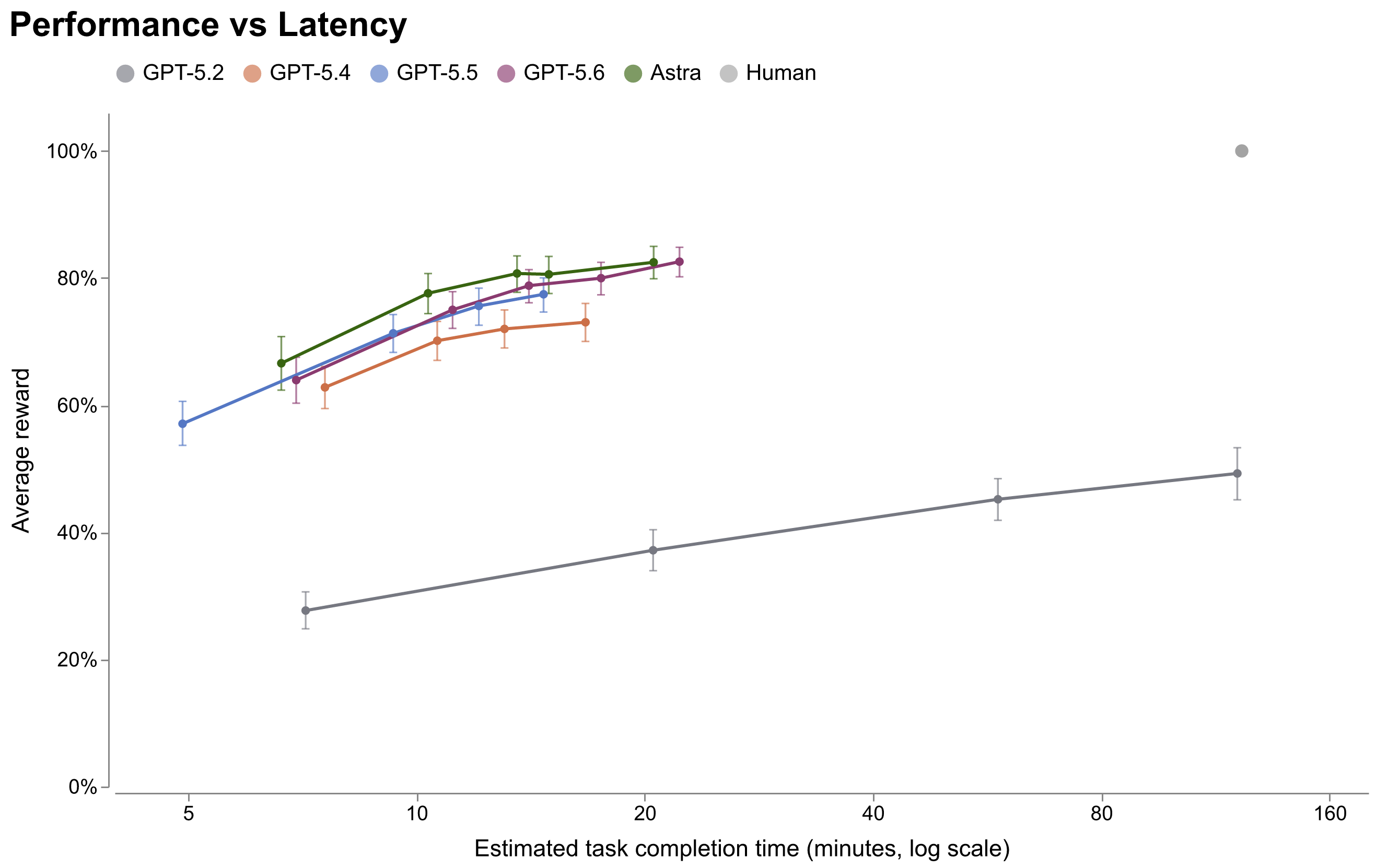}
        
    \end{subfigure}
    \hfill
    \begin{subfigure}{0.48\textwidth}
        \centering
        \includegraphics[width=\linewidth]{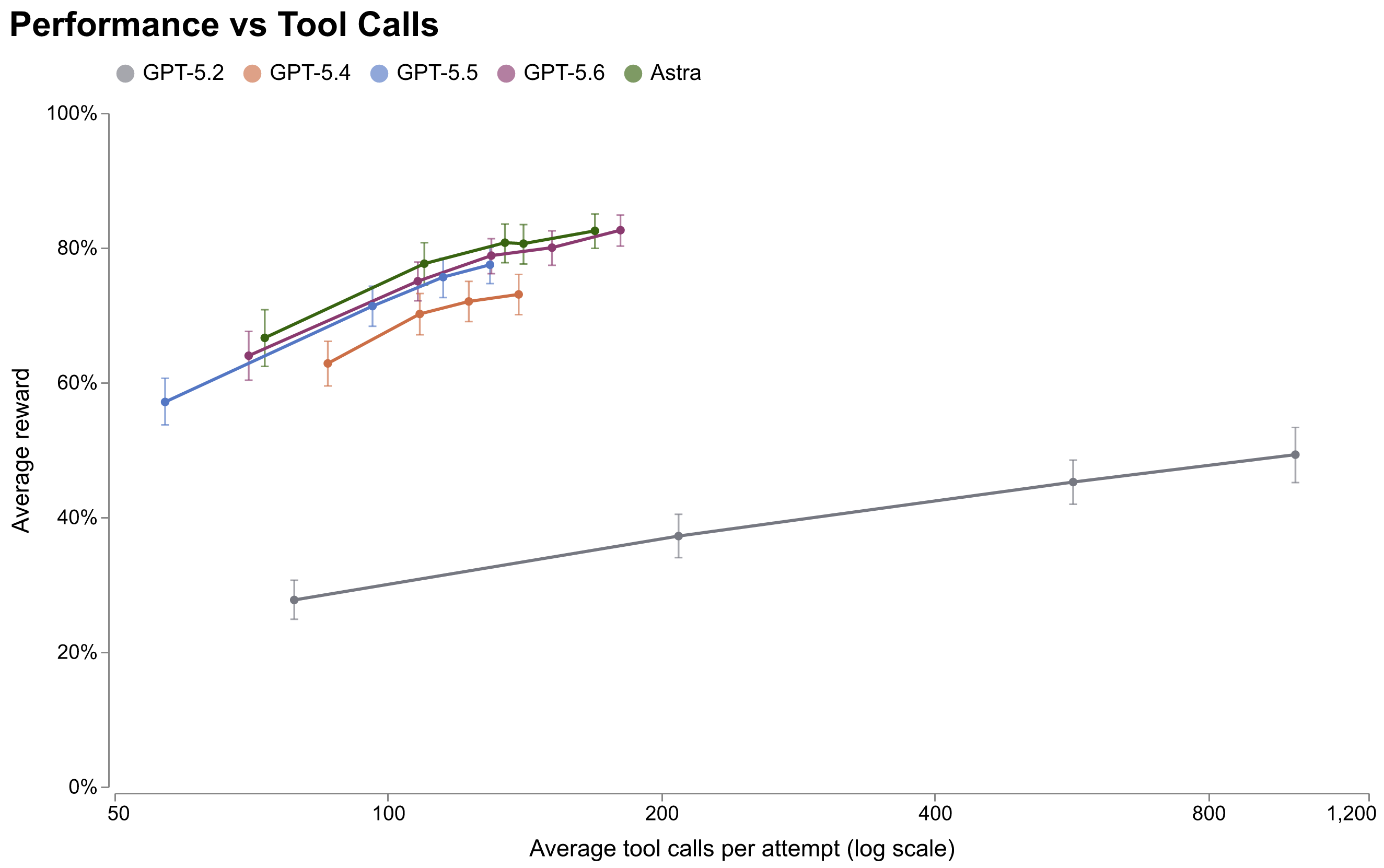}
        
    \end{subfigure}

    \vspace{0.4cm}

    \begin{subfigure}{0.48\textwidth}
        \centering
        \includegraphics[width=\linewidth]{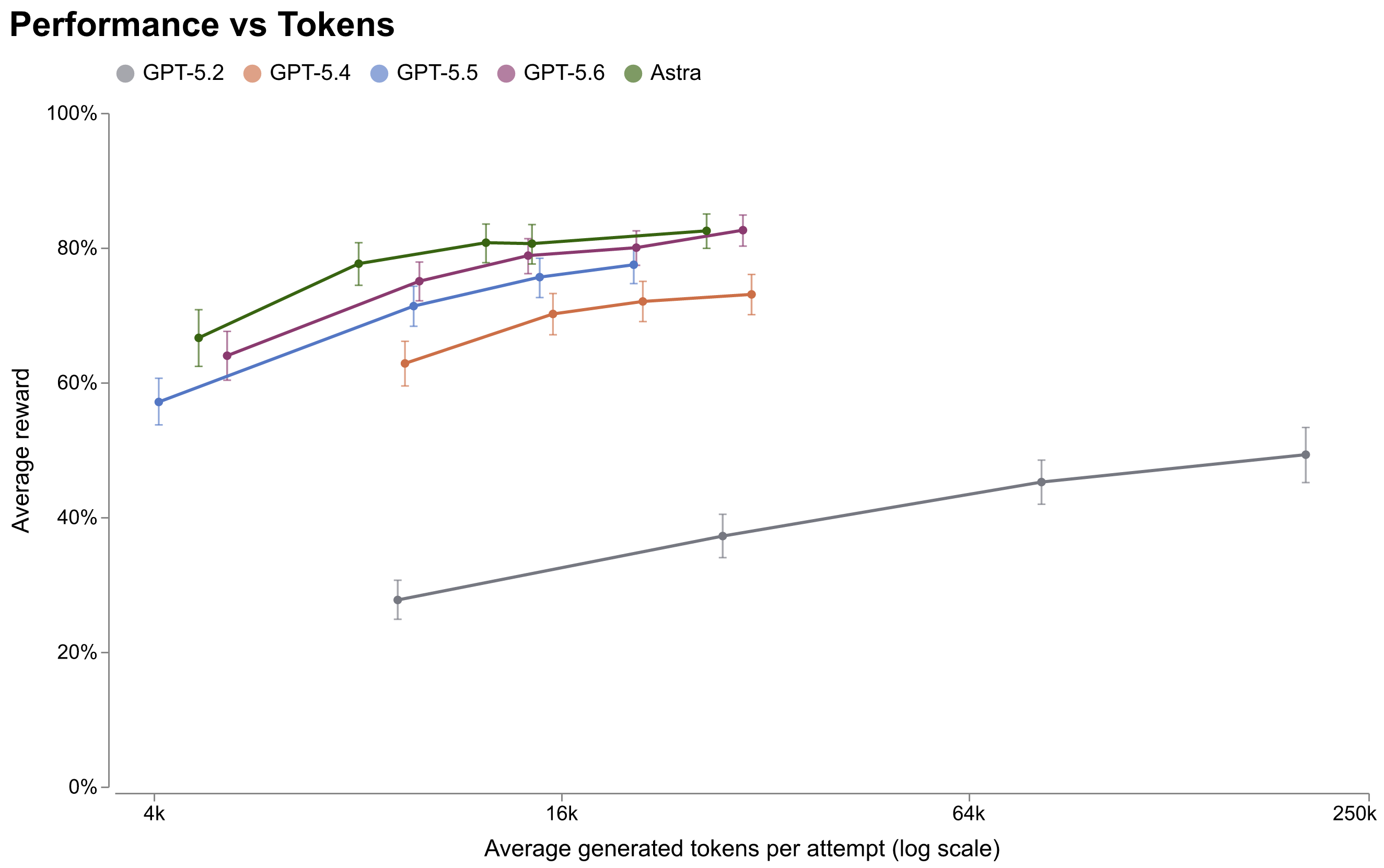}
        
    \end{subfigure}
    \hfill
    \begin{subfigure}{0.48\textwidth}
        \centering
        \includegraphics[width=\linewidth]{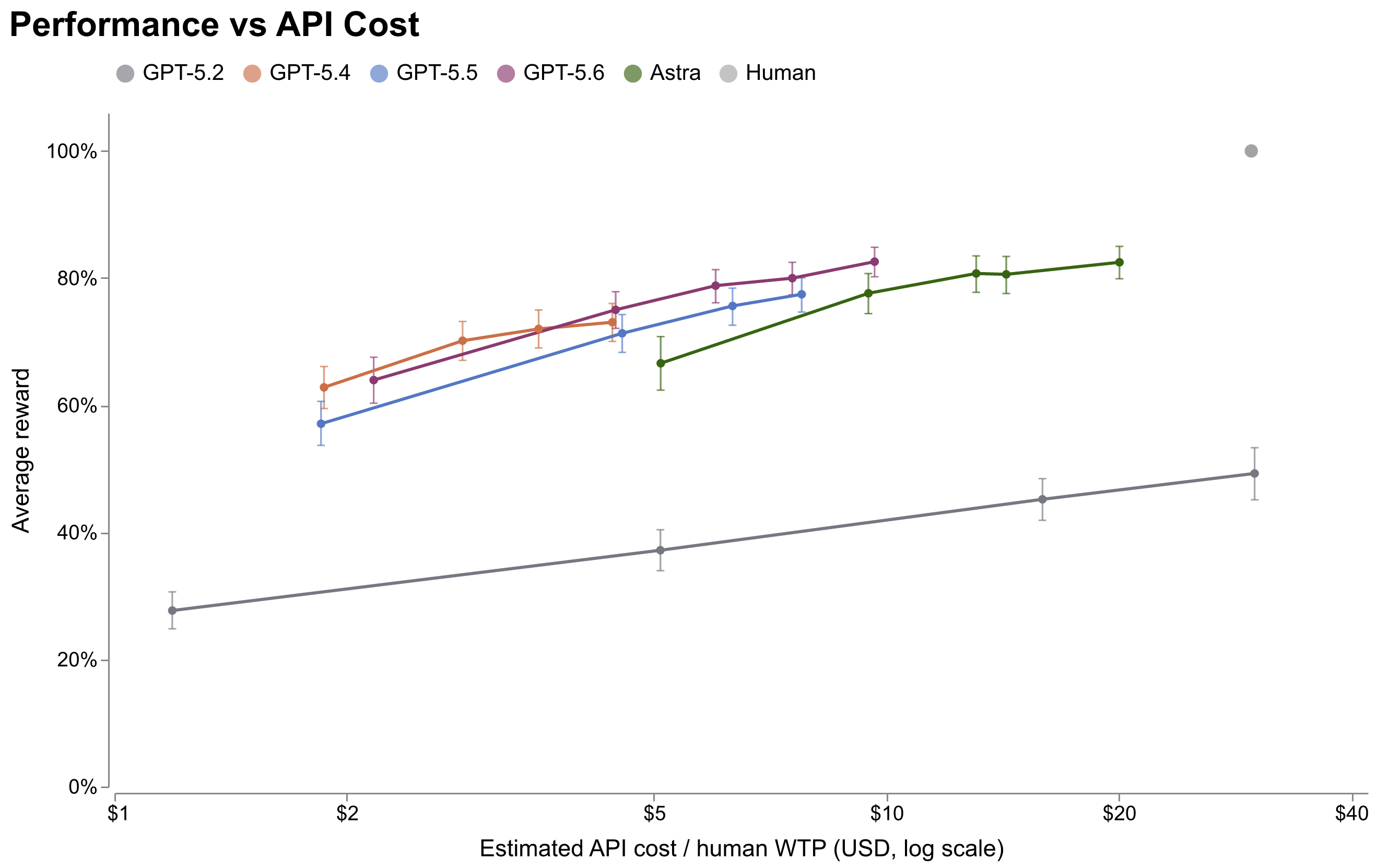}
        
    \end{subfigure}

    \caption{Clockwise from top left: performance versus average observed rollout wall time, average tool calls per rollout, average API cost, and average tokens per rollout. Points show mean reward for each model setting, with an 8-hour timeout per rollout.}
    \label{fig:oss_results}
\end{figure}

The plots show 95\% confidence intervals obtained by bootstrapping (resampling with replacement the available automated grader scores for each sample, computing the mean per sample, and averaging across all samples or for the specified model).  

\subsection{Common Failure Modes}

\begin{figure}[H]
    \centering
    \includegraphics[width=0.75\linewidth]{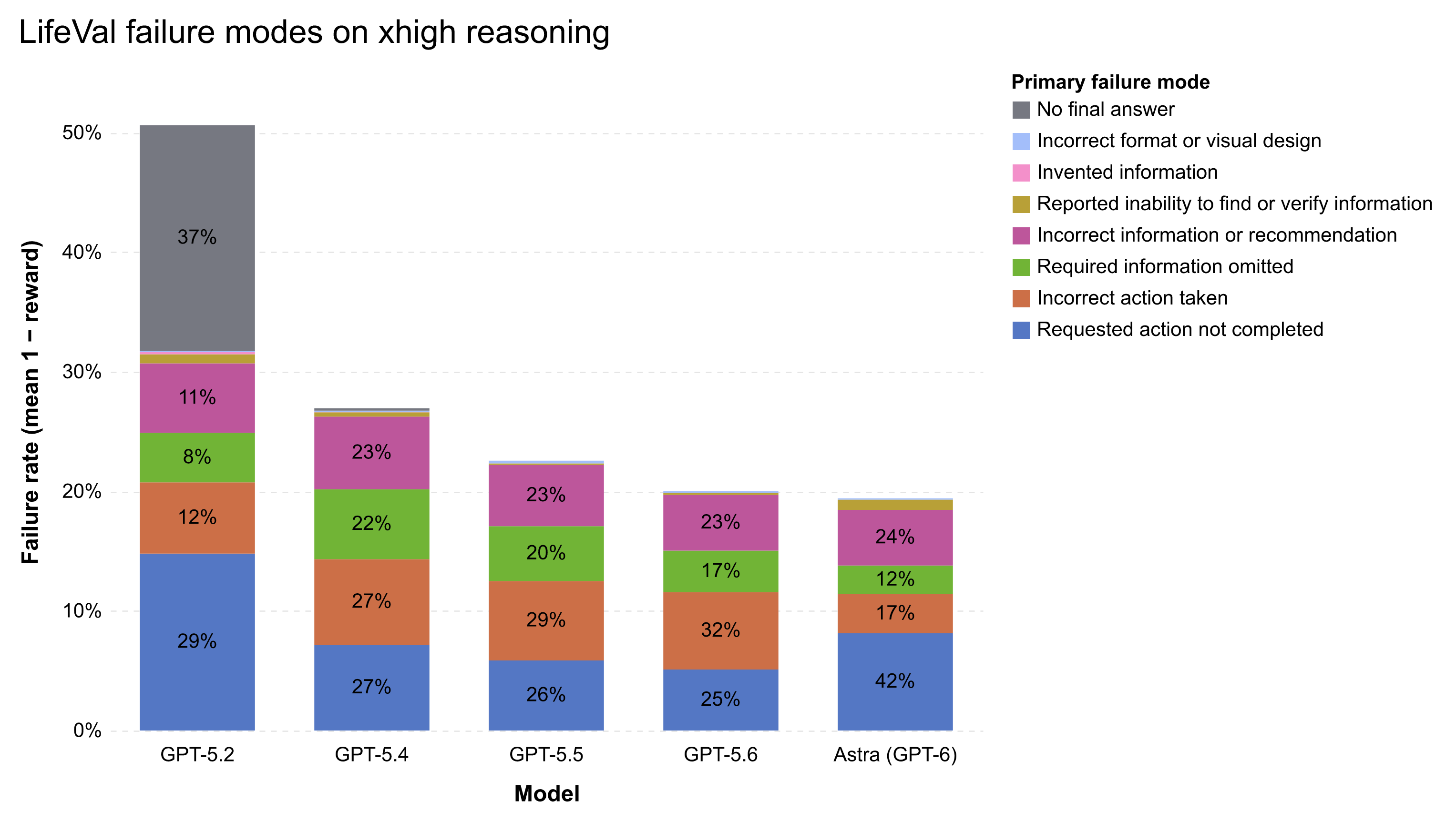}
    \caption{Failure mode profiles by model. Each bar’s total height is the model’s overall failure rate, computed as the average value of \(1 - \text{reward}\) across evaluated samples.}
    \label{fig:failure_modes}
\end{figure}

The models exhibit similar failure-mode profiles but we see trends over time. GPT-5.2 is the only model to run out of tokens trying to complete the tasks, which results in the "no final answer" failure. Since 5.2, improved training on computer use, browser use, and connectors has resulted in higher rewards overall. GPT-6 Astra is our most aligned model yet, and we see an increase in failure modes related to decision boundaries and alignment like ``Required action not completed'' and ``Reported inability to find or verify information.'' The most stable failure mode across models is, ``Incorrect information or recommendation.'' Models often miss key information that is located in personal conversations. 

\section{Limitations}
\label{sec:limitations}

LifeVal has the following limitations:

First, because LifeVal draws on ATUS and MTUS, it may miss newer tasks that do not appear in historical time-use surveys. ATUS ends in 2021, and the most recent survey data from MTUS, which is a database rather than a single survey, is also from 2021. Also, our willingness-to-pay survey was US-centered and conducted in English, so it may under-represent tasks that are common outside the US or in communities that are weakly represented in demographic survey panels.

Second, willingness-to-pay estimates are only a proxy for digital superassistant value. They are likely better than raw time spent for consumer tasks, but they still depend on self-reporting, survey framing, and respondents' imagination of what a ``digital superassistant'' can do. We cross-checked our weights against BLS consumer spending and other related measures, but the final benchmark is still anchored to survey answers rather than revealed long term behavior.

A final limitation is timelessness. Although we direct most web interactions through our RLEs rather than live websites, tasks that depend on time-sensitive information or browsing hit a web cache that is not frozen and can therefore change across runs. This source of variation is substantially reduced by the RLE setup, but not fully removed, and can lead to variations in model scores on tasks that rely on browsing.

\section{Conclusion}
\label{sec:conclusion}
LifeVal introduces the first benchmark we've built specifically to measure AI models' capabilities on consumer-oriented tasks, leveraging personalized ChatGPT data to evaluate performance realistically and accurately. Key contributions of this work include:

\begin{itemize}
    \item \textbf{Task Distribution:} A set of tasks derived from real-world consumer activities and prioritized by consumer-reported value (using a new consumer survey), ensuring relevance and utility. 
    \item \textbf{Dataset:} A novel dataset of tasks, RLEs, rubrics, and personalized, user-specific context drawn directly from the task writers' ChatGPT histories.  
\end{itemize}

LifeVal provides a structured and gradable framework to measure AI models' potential to act as genuinely useful personalized assistants.

\section{Acknowledgments}
We thank Alexandra Barr, Ally Bennett, Andy Applebaum, Annie Lee, April Won, Dana Palmie, Daniel Fradin, David Li, Erik Ritter, Fidji Simo, Grace Kim, Henry Baer, James Lennon, Kendra Rimbauch, Kirsty Le, Laurance Fauconnet, Leon Maksin, Mark Chen, Marwan Aljubeh, Michael Sharman, Michele Wang, Michelle Pokrass, Minnia Feng, Natalie Kim, Nick Otis, Nick Turley, Tom Stasi, Tony Song, Ugurcan Turkdogan, Vinnie Monaco, Wyatt Thompson, Xiaolei Zhu, and Yiheng Xu. We'd especially like to thank the experts who contributed their time and expertise to LifeVal, without whom this work would not have been possible.

\bibliographystyle{plainnat}
\bibliography{references}

@incollection{acemoglu2011skills,
title={{Skills, tasks and technologies: Implications for employment and earnings}},
author={Acemoglu, Daron and Autor, David},
booktitle={Handbook of labor economics},
volume={4},
pages={1043--1171},
year={2011},
publisher={Elsevier}
}

@article{acemoglu2019automation,title={{Automation and new tasks: How technology displaces and reinstates labor}},author={Acemoglu, Daron and Restrepo, Pascual},journal={Journal of economic perspectives},
volume={33},
number={2},
pages={3--30},
year={2019},
publisher={American Economic Association 2014 Broadway, Suite 305, Nashville, TN 37203-2418}
}

@article{autor2003skill,
title={{The skill content of recent technological change: An empirical exploration}},
author={Autor, David H and Levy, Frank and Murnane, Richard J},
journal={The Quarterly journal of economics},
volume={118},
number={4},
pages={1279--1333},
year={2003},
publisher={MIT Press}
}

@article{brynjolfsson2003consumer,
title={{Consumer surplus in the digital economy: Estimating the value of increased product variety at online booksellers}},
author={Brynjolfsson, Erik and Hu, Yu and Smith, Michael D},
journal={Management science},
volume={49},
number={11},
pages={1580--1596},
year={2003},
publisher={INFORMS}
}

@article{dolfen2023assessing,
title={{Assessing the gains from e-commerce}},
author={Dolfen, Paul and Einav, Liran and Klenow, Peter J and Klopack, Benjamin and Levin, Jonathan D and Levin, Larry and Best, Wayne},
journal={American Economic Journal: Macroeconomics},
volume={15},
number={1},
pages={342--370},
year={2023},
publisher={American Economic Association 2014 Broadway, Suite 305, Nashville, TN 37203-2425}
}

@article{gollin2002role,
title={{The role of agriculture in development}},
author={Gollin, Douglas and Parente, Stephen and Rogerson, Richard},
journal={American economic review},
volume={92},
number={2},
pages={160--164},
year={2002},
publisher={American Economic Association}
}

@article{greenwood2005engines,
title={{Engines of liberation}},
author={Greenwood, Jeremy and Seshadri, Ananth and Yorukoglu, Mehmet},
journal={The Review of economic studies},
volume={72},
number={1},
pages={109--133},
year={2005},
publisher={Wiley-Blackwell}
}

@article{herrendorf2013two,
title={{Two perspectives on preferences and structural transformation}},
author={Herrendorf, Berthold and Rogerson, Richard and Valentinyi, Akos},
journal={American Economic Review},
volume={103},
number={7},
pages={2752--2789},
year={2013},
publisher={American Economic Association}
}

@article{ramey2009time,
title={{Time spent in home production in the twentieth-century United States: New estimates from old data}},
author={Ramey, Valerie A},
journal={The Journal of Economic History},
volume={69},
number={1},
pages={1--47},
year={2009},
publisher={Cambridge University Press}
}

@article{sinai2004geography,
title={{Geography and the Internet: Is the Internet a Substitute or a Complement for Cities?}},
author={Sinai, Todd and Waldfogel, Joel},
journal={Journal of Urban Economics},
volume={56},
number={1},
pages={1--24},
year={2004},
publisher={Elsevier}
}

@article{aguiar2007measuring,
title={{Measuring trends in leisure: the allocation of time over five decades}},
author={Aguiar, Mark and Hurst, Erik},
journal={The quarterly journal of economics},
volume={122},
number={3},
pages={969--1006},
year={2007},
publisher={MIT Press}
}

@article{bhargava2015psychological,
title={{Psychological frictions and the incomplete take-up of social benefits: Evidence from an IRS field experiment}},
author={Bhargava, Saurabh and Manoli, Dayanand},
journal={American Economic Review},
volume={105},
number={11},
pages={3489--3529},
year={2015},
publisher={American Economic Association 2014 Broadway, Suite 305, Nashville, TN 37203}
}

@misc{currie2004take,
title={{The take up of social benefits}},
author={Currie, Janet},
year={2004},
publisher={National Bureau of Economic Research Cambridge, Mass., USA}
}

@article{finkelstein2019take,
title={{Take-up and targeting: Experimental evidence from SNAP}},
author={Finkelstein, Amy and Notowidigdo, Matthew J},
journal={The Quarterly Journal of Economics},
volume={134},
number={3},
pages={1505--1556},
year={2019},
publisher={Oxford University Press}
}

@book{herd2019administrative,
title={{Administrative burden: Policymaking by other means}},
author={Herd, Pamela and Moynihan, Donald P},
year={2019},
publisher={Russell Sage Foundation}
}

@misc{osworld2024,
      title={{OSWorld: Benchmarking Multimodal Agents for Open-Ended Tasks in Real Computer Environments}}, 
      author={Tianbao Xie and Danyang Zhang and Jixuan Chen and Xiaochuan Li and Siheng Zhao and Ruisheng Cao and Toh Jing Hua and Zhoujun Cheng and Dongchan Shin and Fangyu Lei and Yitao Liu and Yiheng Xu and Shuyan Zhou and Silvio Savarese and Caiming Xiong and Victor Zhong and Tao Yu},
      year={2024},
      eprint={2404.07972},
      archivePrefix={arXiv},
      primaryClass={cs.AI},
      url={https://arxiv.org/abs/2404.07972}, 
}

@misc{webarena2024,
      title={{WebArena: A Realistic Web Environment for Building Autonomous Agents}}, 
      author={Shuyan Zhou and Frank F. Xu and Hao Zhu and Xuhui Zhou and Robert Lo and Abishek Sridhar and Xianyi Cheng and Tianyue Ou and Yonatan Bisk and Daniel Fried and Uri Alon and Graham Neubig},
      year={2024},
      eprint={2307.13854},
      archivePrefix={arXiv},
      primaryClass={cs.AI},
      url={https://arxiv.org/abs/2307.13854}, 
}

@misc{osworldverified2025,
  title        = {{Introducing OSWorld-Verified}},
  author       = {{XLANG Lab}},
  howpublished = {\url{https://xlang.ai/blog/osworld-verified}},
  year         = {2025},
  note         = {Accessed 2026-04}
}

@misc{webvoyager2024,
  title        = {{WebVoyager: Building an End-to-End Web Agent with Large Multimodal Models}},
  author       = {He, Hongliang and Yao, Wenlin and Ma, Kaixin and Yu, Wenhao and Dai, Yong and Zhang, Hongming and Lan, Zhenzhong and Yu, Dong},
  howpublished = {arXiv preprint arXiv:2401.13919},
  year         = {2024}
}

@misc{openai_gpt54_2026,
  title        = {{Introducing GPT-5.4}},
  author       = {{OpenAI}},
  howpublished = {\url{https://openai.com/index/introducing-gpt-5-4/}},
  year         = {2026},
  note         = {Accessed 2026-04}
}

@misc{rdi_benchmark_hacking_2026,
  title        = {{How We Broke Top AI Agent Benchmarks: And What Comes Next}},
  author       = {Wang, Hao and Mang, Qiuyang and Cheung, Alvin and Sen, Koushik and Song, Dawn},
  howpublished = {\url{https://rdi.berkeley.edu/blog/trustworthy-benchmarks-cont/}},
  year         = {2026},
  note         = {Accessed 2026-04}
}

@misc{permabenchmarking2026,
      title={{PERMA: Benchmarking Personalized Memory Agents via Event-Driven Preference and Realistic Task Environments}}, 
      author={Shuochen Liu and Junyi Zhu and Long Shu and Junda Lin and Yuhao Chen and Haotian Zhang and Chao Zhang and Derong Xu and Jia Li and Bo Tang and Zhiyu Li and Feiyu Xiong and Enhong Chen and Tong Xu},
      year={2026},
      eprint={2603.23231},
      archivePrefix={arXiv},
      primaryClass={cs.AI},
      url={https://arxiv.org/abs/2603.23231}, 
}

@misc{personalizeddeepresearch2026,
      title={{Towards Personalized Deep Research: Benchmarks and Evaluations}}, 
      author={Yuan Liang and Jiaxian Li and Yuqing Wang and Piaohong Wang and Motong Tian and Pai Liu and Shuofei Qiao and Runnan Fang and He Zhu and Ge Zhang and Minghao Liu and Yuchen Eleanor Jiang and Ningyu Zhang and Wangchunshu Zhou},
      year={2026},
      eprint={2509.25106},
      archivePrefix={arXiv},
      primaryClass={cs.CL},
      url={https://arxiv.org/abs/2509.25106}, 
}

@misc{surveypersonalization2025,
      title={{A Survey of Personalization: From RAG to Agent}}, 
      author={Xiaopeng Li and Pengyue Jia and Derong Xu and Yi Wen and Yingyi Zhang and Wenlin Zhang and Wanyu Wang and Yichao Wang and Zhaocheng Du and Xiangyang Li and Yong Liu and Huifeng Guo and Ruiming Tang and Xiangyu Zhao},
      year={2025},
      eprint={2504.10147},
      archivePrefix={arXiv},
      primaryClass={cs.IR},
      url={https://arxiv.org/abs/2504.10147}, 
}

@techreport{bls2024atus,
  author = {{Bureau of Labor Statistics}},
  title = {{American Time Use Survey -- 2024 Results}},
  institution = {U.S. Department of Labor},
  year = {2025},
  url = {https://www.bls.gov/news.release/pdf/atus.pdf},
  note = {Accessed: April 16, 2026}
}

@misc{mtus2020,
  author = {Fisher, Kimberly and Gershuny, Jonathan and Flood, Sarah M. and
            Lamote, Juana and Sayer, Liana C. and Backman, Daniel and
            Breton, Etienne and Richards, Stephanie},
  title = {{Multinational Time Use Study Extract System: Version 1.5 [dataset]}},
  year = {2025},
  howpublished = {Minneapolis, MN: IPUMS},
  doi = {10.18128/D062.V1.5},
  url = {https://doi.org/10.18128/D062.V1.5}
}

@misc{BLS_CE_2024,
  author = {{U.S. Bureau of Labor Statistics}},
  title = {{Consumer Expenditures—2024}},
  year = {2025},
  url = {https://www.bls.gov/news.release/pdf/cesan.pdf},
  note = {Accessed: April 16, 2026}
}

@misc{onet2026,
  author = {{National Center for O*NET Development}},
  title = {{O*NET OnLine}},
  url = {https://www.onetonline.org/},
  year = {2026},
  note = {Accessed: April 16, 2026}
}

@misc{ATUSWB2021,
  author = {{Bureau of Labor Statistics}},
  title = {{American Time Use Survey -- 2021 Well-Being Module Microdata Files}},
  year = {2022},
  howpublished = {Machine-readable data file},
  url = {https://www.bls.gov/tus/data/wbdatafiles-2021.htm},
  note = {Accessed: 2026-04-17}
}

@inproceedings{zhao2025prefeval,
  title={{Do LLMs Recognize Your Preferences? Evaluating Personalized Preference Following in LLMs}},
  author={Zhao, Siyan and Hong, Mingyi and Liu, Yang and Hazarika, Devamanyu and Lin, Kaixiang},
  booktitle={International Conference on Learning Representations (ICLR)},
  year={2025},
  url={https://www.amazon.science/publications/do-llms-recognize-your-preferences-evaluating-personalized-preference-following-in-llms}
}

@article{patwardhan2025gdpval,
  title={{GDPval: Evaluating AI Model Performance on Real-World Economically Valuable Tasks}},
  author={Patwardhan, Tejal and Dias, Rachel and Proehl, Elizabeth and Tworek, Jerry and others},
  journal={arXiv preprint arXiv:2510.04374},
  year={2025},
  url={https://arxiv.org/abs/2510.04374}
}

@techreport{census2021population,
  author = {{U.S. Census Bureau}},
  title = {{Annual Estimates of the Resident Population for the
           United States, Regions, States, the District of Columbia,
           and Puerto Rico: April 1, 2020 to July 1, 2021}},
  institution = {U.S. Department of Commerce},
  year = {2021},
  number = {NST-EST2021-POP},
  url = {https://www2.census.gov/programs-surveys/popest/tables/2020-2021/state/totals/NST-EST2021-POP.xlsx},
  note = {Vintage 2021. Accessed: April 16, 2026}
}

@techreport{cpsasec2025,
  author = {{U.S. Census Bureau}},
  title = {{Current Population Survey: 2025 Annual Social and
           Economic Supplement ({CPS-ASEC})}},
  institution = {U.S. Department of Commerce},
  type = {Public-use microdata},
  year = {2025},
  url = {https://www.census.gov/data/datasets/2025/demo/cps/cps-asec-2025.html},
  note = {Accessed: April 16, 2026}
}

@inproceedings{kim2026persona2web,
  title     = {{Persona2Web}: Benchmarking Personalized Web Agents
               for Contextual Reasoning with User History},
  author    = {Kim, Serin and Lee, Sangam and Lee, Dongha},
  booktitle = {Proceedings of the 43rd International Conference
               on Machine Learning},
  series    = {Proceedings of Machine Learning Research},
  volume    = {306},
  pages     = {58276--58303},
  year      = {2026},
  publisher = {PMLR},
  url       = {https://proceedings.mlr.press/v306/kim26ap.html}
}

@inproceedings{cai2025personalwab,
  title     = {Large Language Models Empowered Personalized Web Agents},
  author    = {Cai, Hongru and Li, Yongqi and Wang, Wenjie
               and Zhu, Fengbin and Shen, Xiaoyu and Li, Wenjie
               and Chua, Tat-Seng},
  booktitle = {Proceedings of the ACM on Web Conference 2025},
  series    = {WWW '25},
  pages     = {198--215},
  year      = {2025},
  publisher = {Association for Computing Machinery},
  doi       = {10.1145/3696410.3714842},
  url       = {https://doi.org/10.1145/3696410.3714842}
}
\appendix

\section{Appendix}

\subsection{ATUS Data Cleaning Procedure}
\label{app:datacleaning}
We performed the following data cleaning.
\begin{itemize}
    \item \textbf{Merged related subcategories:} We collapsed related subcategories like traveling to various places, watching various sports, and waiting activities into unified categories for ``transportation,'' ``watching sports'' and ``waiting.''
    \item \textbf{Merged closely related tasks:} We consolidated very similar tasks like ``medical care provided to adults'' and ``medical care provided to children'' into one category and combined financial management tasks regardless of whether assistance is provided to others or for oneself. 
    \item \textbf{Removed long-tail categories:} We excluded work-related activities (already covered in GDPval) and unspecified categories (``not elsewhere classified'') from our analysis. 
\end{itemize}

\subsection{Consumer willingness-to-pay survey}
\label{app:survey}

Recruitment targeted U.S. population distributions for age, gender, ethnicity, and income. 3,837 respondents  completed the survey. Screening comprised comprehension questions assessing respondents’ understanding of the assistant’s capabilities and a simple attention check. Respondents were retained only if they answered all five checks correctly, yielding 1,826 retained respondents and 2,011 exclusions.

Each respondent provided valuations for a subset of up to 30 activities from the cleansed ATUS/MTUS task distribution. Each activity-level estimate was calculated from the retained respondents who provided a valuation for that activity. For each assigned activity, respondents were asked the following valuation question:
\begin{quote}

Imagine you have access to a perfect digital superassistant.

The digital superassistant provides perfect advice (e.g., explain how to fix your washing machine, give the world’s best personalized parenting tips, create a custom LSAT study guide), and when possible, automates digital tasks on your behalf (e.g., buying your preferred clothing online, filling out an online tax return perfectly on your behalf, knowing what takeout dinner you want and ordering it via delivery).
\begin{itemize}
    \item Any money that the digital superassistant spends comes from your wallet.
    \item Assume the digital superassistant is perfect at performing these tasks–so for the purposes of this exercise, assume that it knows your preferences perfectly and does not make any mistakes.
\end{itemize}

Consider the entirety of last year (2025). What is the most you would have been willing to pay, as the sum total across the whole year, in \$USD, for the digital superassistant to assist on this task? Put 0 if you would not pay to use the digital superassistant.

\end{quote}
We estimated annual willingness to pay for each activity using a 95th-percentile winsorized mean. Specifically, valuations above the 95th percentile for the activity were replaced by that percentile, and the resulting values were averaged. Percentiles were calculated by linear interpolation. Zero valuations were included, while missing responses were excluded from the denominator for the activity.

We converted these estimates into task collection targets proportional to estimated willingness to pay. Each activity’s share of total estimated willingness to pay across eligible activities was multiplied by a target of 800 tasks and rounded down to the nearest integer. After rounding, 148 activity categories had positive allocations, corresponding to 702 tasks representing more than \$1.7T in annual willingness to pay for US consumers. 

\subsubsection{Validating willingness-to-pay survey results}
\label{app:surveyvalidation}

As a robustness check, we compared our willingness-to-pay rankings against several external proxies for task importance: the BLS Consumer Expenditure Survey, O*NET, and the ATUS Wellbeing Survey \citep{BLS_CE_2024, onet2026, ATUSWB2021}. 

For each ATUS Tier 3 activity in our distribution, we mapped a corresponding spending category, occupation, and ATUS wellbeing activity where possible. For example, the activity ``using legal services'' was mapped to the spending category ``legal fees'' and the occupation ``Lawyers.'' We then compared our survey-reported willingness to pay against external measures of willingness to pay, including average household spending, spending participation rates, occupational wages, and reported emotions during the activity.

These external measures were only moderately correlated with our survey-based willingness-to-pay rankings ($R^2 \approx 0.4$). We view this as evidence that willingness to pay captures something related to, but distinct from, standard proxies such as spending, wages, or experienced affect. In particular, we believe the differences arise for several reasons:
\begin{itemize}
    \item Consumers may place high value on tasks that have important long-run consequences even when they involve relatively little direct spending (e.g., financial management).
    \item Occupational wages reflect labor market conditions, not just value to consumers.
    \item Consumer expenditure categories combine goods and services, whereas willingness to pay for AI assistance is primarily thought of as a service.
    \item Negative or positive emotions during an activity do not consistently translate into willingness to pay for that activity.
\end{itemize}

\subsection{Grading}
\label{app:grading}
Our main grader for the above results was GPT-5.6 Sol with xhigh reasoning. To test whether our autograder aligned with human judgments, we also asked human trainers to grade the model’s rollouts using the task rubrics. Seventeen trainers graded one GPT-5.5 rollout for each task in an n=30 subset, and we found that the mean rubric grade was 83.61\% for humans and 84.20\% for GPT-5.6. Across the 30 rollouts, the Pearson correlation between the mean human grade and the GPT-5.6 grade was $r = 0.864$.

\begin{table}[H]
  \centering
  \small
  \label{tab:lifeval_human_autograder}
  \begin{tabular}{lc}
    \toprule
    Grader & Mean rubric grade \\
    \midrule
    Human   & 83.61\% \\
    GPT-5.6 & 84.20\% \\
    \bottomrule
  \end{tabular}
\end{table}

\section{{Personalization}}

To assess the importance of personalization in LifeVal tasks, we evaluated the tasks under four conditions: baseline; past conversations and memories removed; additional task context removed; and additional task context synthetically inserted into past conversations rather than provided directly. We used the same model and varied reasoning effort across all conditions.

Removing additional task context substantially reduced performance at every reasoning effort, while removing past conversations and memories had a much smaller effect. Inserting facts from the additional task context into past conversations reduced performance by approximately as much as removing the context entirely. These results suggest two conclusions:
\begin{enumerate}
    \item Additional task context contains much of the personal information needed to complete the tasks.
    \item The model has difficulty finding and using relevant information in past conversations.
\end{enumerate}

To investigate these findings further, we examined performance on personalized rubric criteria: criteria requiring information from memories, past conversations, or additional task context. When creating their rubrics, trainers identified the where the information needed to satisfy each criterion was located in the task (e.g. the reference files, their past conversations, additional task context, available via browsing, etc.). We selected criteria where trainers identified exactly one of these three sources and compared their win rates. 

\begin{itemize}
    \item Past conversations: average win rate of 33.33\%.
    \item Memories: average win rate of 47.8\%.
    \item Additional task context: average win rate of 55.8\%, falling to 41.5\% when the information was synthetically inserted into past conversations.
\end{itemize}

For criteria attributed exclusively to past conversations on which the model achieved a 100\% win rate, we inspected the corresponding rollouts to determine:
\begin{enumerate}
    \item Whether the model accessed past conversations.
    \item Whether it found the relevant information.
    \item Whether it used that information correctly.
    \item Whether the information was also available elsewhere in the task or could be inferred from conversation titles alone.
\end{enumerate}

For every criterion in this reviewed subset, the information needed to satisfy it was also available elsewhere in the task or could be inferred from conversation titles alone. Thus, success on these criteria did not establish that the model retrieved the relevant facts from conversation contents. We observed the model searching conversation history, including in unsuccessful attempts, but its searches often relied on keywords and topical cues. For example:

\begin{lstlisting}[
    language=bash,
    basicstyle=\ttfamily\footnotesize,
    columns=fullflexible,
    keepspaces=true,
    showstringspaces=false,
    breaklines=true,
    breakatwhitespace=false,
    linewidth=\linewidth
]
grep -Rni -E \
  'GAship|graduate assistant|full.?time|student loan|LING 800|dissertation credits|qualifying paper|registration' \
  /home/reference_files /home/conversation_history \
  2>/dev/null | head -300
\end{lstlisting}

\subsection{Expert Sourcing, Instructions, and Compensation}
\label{app:contributors}

Experts were recruited from existing pools of generalist task writers, supplemented with targeted outreach for niche categories (e.g. ADU setup, mold remediation) via blogs and social media channels.

Participation was voluntary, and contributors could withdraw at any time. Contributors were compensated as professional contractors for their time and expertise. 

\subsection{ATUS/MTUS Task Categories}
\label{app:task_categories}
% Requires: \usepackage{booktabs}

\begin{table}[t]
\centering
\scriptsize
\setlength{\tabcolsep}{3pt}
\renewcommand{\arraystretch}{0.98}
\begin{tabular}{@{}>{\raggedright\arraybackslash}p{0.22\linewidth}>{\raggedright\arraybackslash}p{0.42\linewidth}>{\raggedleft\arraybackslash}p{0.15\linewidth}>{\raggedleft\arraybackslash}p{0.15\linewidth}@{}}
\toprule
Tier 1 activity category & Sample of Tier 3 activities & Annual WTP per respondent (US\$, sum) & US annual WTP (US\$ billions) \\
\midrule
Household activities and services & Financial management; Using home maint/repair/decor/construction svcs; Exterior repair, improvements, \& decoration & \$1,811.7 & \$505 \\
Caring for and helping others & Home schooling of children; Looking after adult (as a primary activity); Activities related to child's education, n.e.c.* & \$1,289.4 & \$329 \\
Personal care & Activities rel. to purchasing/selling real estate; Using health and care services outside the home; Banking & \$856.3 & \$239 \\
Education & Research/homework for class for degree, certification, or licensure; Research/homework n.e.c.*; Taking class for degree, certification, or licensure & \$653.7 & \$190 \\
Volunteer activities & Volunteer - Computer use; Volunteer - Teaching, leading, counseling, mentoring; Volunteer - Telephone calls (except hotline counseling) & \$385.5 & \$100 \\
Consumer purchases & Shopping, except groceries, food and gas; Grocery shopping; Comparison shopping & \$266.9 & \$81 \\
Socializing, relaxing and leisure & Writing for personal interest; Hobbies, except arts \& crafts and collecting; Arts and crafts as a hobby & \$282.3 & \$77 \\
Telephone calls & Telephone calls to/from professional or personal care svcs providers; Telephone calls to/from salespeople; Telephone calls to/from household services providers & \$233.8 & \$64 \\
Professional and personal care services & Job search activities; Job interviewing & \$159.6 & \$47 \\
Government services and civic obligations & Obtaining licenses \& paying fines, fees, taxes; Using social services; Civic obligations \& participation & \$158.7 & \$46 \\
Sports, exercise and recreation & Weightlifting/strength training; Working out, unspecified; Playing baseball & \$135.7 & \$35 \\
Traveling & Traveling; Travel as a form of entertainment & \$104.2 & \$28 \\
Eating and drinking & Eating and drinking & \$13.2 & \$3 \\
Religious and spiritual activities & Religious education activities & \$9.3 & \$2 \\
\midrule
\multicolumn{3}{@{}l}{\textbf{Total across listed categories}} & \textbf{\$1,746} \\
\bottomrule
\end{tabular}
\caption{ATUS/MTUS Tier 1 activity categories, sample Tier 3 activities, and summed willingness-to-pay estimates. The annual WTP column sums activity-level means from our demographic survey; the U.S. annual WTP column sums national estimates.}
\label{tab:atus-categories}
\end{table}

\FloatBarrier

\end{document}